%% file: main.tex
\documentclass[runningheads]{llncs}

\usepackage{eccv}

\usepackage{eccvabbrv}

\usepackage{graphicx}
\usepackage{booktabs}

\usepackage{url}

\usepackage{amsfonts}
\usepackage{amssymb}
\usepackage{multirow}
\usepackage{makecell}
\usepackage{booktabs}

\usepackage{bm}
\usepackage{color}
\usepackage{pifont}

\usepackage{algorithmicx}  
\usepackage{algpseudocode} 

\usepackage[accsupp]{axessibility}  

\newcommand{\cmark}{\ding{51}\xspace}%
\newcommand{\xmark}{\ding{55}\xspace}%
\usepackage{hyperref}

\usepackage{orcidlink}

\begin{document}

\title{LayerDiff: Exploring Text-guided Multi-layered Composable Image Synthesis via Layer-Collaborative Diffusion Model} 



\author{Runhui Huang\inst{1} \and
Kaixin Cai\inst{1}  \and
Jianhua Han \inst{2}  \and
Xiaodan Liang \inst{1}
Renjing Pei \inst{2} \and
Guansong Lu \inst{2} \and
Songcen Xu \inst{2} \and
Wei Zhang \inst{2}  \and
Hang Xu \inst{2} }


\institute{Sun Yat-sen University \and Huawei Noah's Ark Lab}

\maketitle


\input{sec/0_abstract}    
\input{sec/1_intro}

\input{sec/2_related_work}
\input{sec/3_method}
\input{sec/4_experiment}
\input{sec/5_conclusion}


%
%
\bibliographystyle{splncs04}
\bibliography{main}

\clearpage  

\appendix
\input{sec/X_suppl}

\end{document}

%% file: sec/0_abstract.tex
\begin{abstract}
\vspace{-3mm}
Despite the success of generating high-quality images given any text prompts by diffusion-based generative models, prior works directly generate the entire images, but cannot provide object-wise manipulation capability. To support wider real applications like professional graphic design and digital artistry, images are frequently created and manipulated in multiple layers to offer greater flexibility and control. 
Therefore in this paper, we propose a layer-collaborative diffusion model, named \textbf{LayerDiff}, specifically designed for text-guided, multi-layered, composable image synthesis.
The composable image consists of a background layer, a set of foreground layers, and associated mask layers for each foreground element. 
To enable this, LayerDiff introduces a layer-based generation paradigm incorporating multiple layer-collaborative attention modules to capture inter-layer patterns.
Specifically, an inter-layer attention module is designed to encourage information exchange and learning between layers, while a text-guided intra-layer attention module incorporates layer-specific prompts to direct the specific-content generation for each layer. A layer-specific prompt-enhanced module better captures detailed textual cues from the global prompt. Additionally, a self-mask guidance sampling strategy further unleashes the model's ability to generate multi-layered images.
We also present a pipeline that integrates existing perceptual and generative models to produce a large dataset of high-quality, text-prompted, multi-layered images. Extensive experiments demonstrate that our LayerDiff model can generate high-quality multi-layered images with performance comparable to conventional whole-image generation methods.  Moreover, LayerDiff enables a broader range of controllable generative applications, including layer-specific image editing and style transfer.
\keywords{Multi-layered Composable Image Synthesis \and Layer-Collaborative Diffusion Model \and Layer-specific Image Editing.}
\end{abstract}

%% file: sec/1_intro.tex
\section{Introduction}
\begin{figure}
\begin{center}
\includegraphics[width=\linewidth]{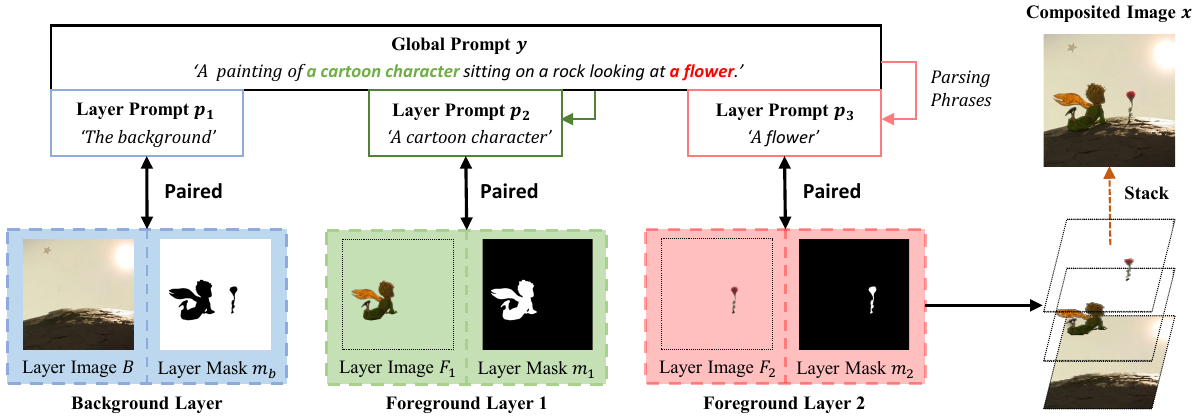}
\label{fig:example_of_data}
\caption{An examples of the multi-layered composable image. The multi-layered composable image includes a background layer, a set of foreground layers and the corresponding layer masks. The layer images, layer masks and the layer prompts with the same color are belonging to the same layer. The text-guided multi-layered composable image synthesis is aimed to generate the layer images and layer masks simultaneously under the guidance of global prompt to control the holistic content, and the layer prompts to control the per layer's content. It's able to composite a whole image by assembling these layers according to the masks. 
}
\label{fig:data_example}
\end{center}
\vspace{-8mm}
\end{figure}
The image generation task from textual descriptions~\cite{nichol2021glide,ramesh2021zero} has emerged as a critical research area in the field of computer vision and machine learning, with broad applications spanning graphic design, advertising, and scientific visualization. Among various paradigms, diffusion-based generative models~\cite{ddpm,song2020denoising,nichol2021glide} have demonstrated superior efficacy in synthesizing high-quality images from text prompts, thereby serving as an invaluable asset for both creative endeavors and practical applications.
Nonetheless, a salient limitation of current methodologies~\cite{goodfellow2020generative,nichol2021improved-ddpm,rombach2022high} is their inherent design to generate only single-layer, monolithic images. Such a limitation substantially restricts their applicability and versatility in scenarios that require a greater degree of control and adaptability, particularly in professional graphic design and digital artistry where layered compositions are frequently indispensable.
In these applications, images are rarely created as a single piece; instead, they are composed and manipulated in multiple layers that allow for iterative refinement, fine-grained adjustment, and object-specific manipulations. 
Therefore, in this paper, we concentrate on building a layer-collaborative diffusion-based framework and achieving the multi-layered, composable image synthesis task.

Several approaches to layered-image synthesis exist, such as the Text2Layer~\cite{zhang2023text2layer}, which employs an autoencoder architecture and a latent diffusion mechanism for dual-layer image synthesis. However, Text2Layer is limited to two-layered compositions, thus not fully capitalizing on the expressive power of multi-layered synthesis. 
Another potential approach can be a post-processing pipeline involving word tokenization~\cite{Honnibal_spaCy}, perceptual segmentation~\cite{kirillov2023segment}, and inpainting techniques~\cite{rombach2022high}. Although functional, this post-processing strategy introduces considerable computational overhead and is prone to error accumulation through multiple processing stages, potentially leading to content or stylistic inconsistencies.
Hence, the central aim of this study is to utilize a unified diffusion model framework for accomplishing end-to-end multi-layer image synthesis.

In this paper, we introduce \textbf{LayerDiff}, a novel layer-collaborative diffusion model tailored for text-guided, multi-layered, and composable image synthesis. In contrast to conventional whole-image generative models, LayerDiff pioneers a new paradigm in layer-based image generation, offering the ability to synthesize images comprising multiple constituent layers. As shown in Fig. \ref{fig:data_example}, the resulting composite image is structured into a background layer, an ensemble of foreground layers, and individual mask layers that define the spatial arrangement of each foreground element.
To achieve fine-grained control over the number of layers and the content within each layer, we introduce layer-specific prompts in addition to the global prompt that governs the overall image content. We present the layer-specific prompt enhancer designed to extract intricate textual features and object relationships from the global prompt. Furthermore, we introduce the layer-collaborative attention block that facilitates inter-layer interaction and layer-specific content modulation through the employment of an inter-layer attention module and a text-guided intra-layer attention module.
To further release the power of our LayerDiff, we propose a self-mask guidance sampling strategy by using the predicted layer masks to enhances the quality of the generated multi-layered images.

To construct a dataset suitable for training LayerDiff, we introduce a carefully designed data acquisition pipeline aimed at generating high-quality, multi-layered composable images. Firstly, we employ state-of-the-art open-vocabulary perception algorithms for object localization to form multiple layers. We use a powerful image captioner to obtain detailed description of whole image and object description for each layer. We employ the inpainting model to fill in the missing portions at the respective locations after extraction.
Comprehensive experiments confirm that our LayerDiff architecture excels in producing high-fidelity multi-layered images, exhibiting performance metrics on par with traditional full-image generative techniques. Additionally, LayerDiff's support for both global and layer-specific prompts expands its utility across a diverse set of controllable generative tasks, such as layer-wise composable image manipulation and style transfer.

To summarize the contribution:
\begin{itemize}
    \item We introduce LayerDiff, a layer-collaborative diffusion model which employs layer-collaborative attention blocks for inter- and intra-layer information exchange. A layer-specific prompt-enhanced module further refines content generation by leveraging global textual cues.
    \item We propose the self-mask guidance sampling further guide the model to generate high-quality multi-layered images by leveraging intermediate layer mask predictions during the sampling process to refine generation results.
    \item We introduce a data construction pipeline that generates multi-layered composable images for LayerDiff, integrating state-of-the-art techniques in image captioning, object localization, segmentation and inpainting.
    \item Our LayerDiff not only generates high-fidelity multi-layered images with performance comparable to traditional whole-image generation methods but also offers versatile control for various generative applications.
\end{itemize}

%% file: sec/2_related_work.tex
\section{Related Work}
\noindent \textbf{Text-to-Image Synthesis.}
The image generation models can be broadly categorized into several architectures, including VAE~\cite{huang2018introvae}, flow-based~\cite{ho2019flow++}, GAN~\cite{brock2018large,kang2023scaling}, autoregressive model~\cite{yu2022parti,wu2022nuwa} and diffusion models~\cite{nichol2021improved-ddpm}. Leveraging the advantages of diffusion models in the field of image generation, many recent works~\cite{imagen,gu2022vector,rombach2022high} have started using large-scale text-image pairs for training models, achieving remarkable performance in the text-to-image generation domain. One of the most representative works is Stable Diffusion~\cite{rombach2022high}, which applies diffusion to the latent image encoded by an autoencoder, and employs a UNet-like architecture to learn the denoising task. It utilizes a text encoder, pretrained with visually aligned text from CLIP~\cite{radford2021learning}, to encode textual descriptions and injects textual guidance into the network in the form of cross-attention. 
Text2Layer~\cite{zhang2023text2layer} proposes a layered image generation model, dividing an image into a foreground image, a foreground mask and a background image and encoding layer information into one latent with layer generation accomplished through a latent diffusion model. 
Distinguishing from the aforementioned models, we firstly introduce the task of multi-layered composable image synthesis. LayerDiff can generate two or more layers and introduces layer-specific prompts to finely control the content generation per layer.

\noindent \textbf{Controllable Image Synthesis.}  
Various approaches have explored to utilize many control signals to guide image generation.
Based on Stable Diffusion, through fine-tuning text embedding, Textual Inversion~\cite{gal2022image} and DreamBooth~\cite{ruiz2023dreambooth} enables the generation of personalized objects in novel scenarios. P2P~\cite{hertz2022prompt2prompt} and PnP~\cite{tumanyan2023plug} introduce attention mechanisms with cross-attention to achieves for text-guided content editing. 
The layout-to-image generation methods~\cite{Xue_2023_CVPR,zheng2023layoutdiffusion} aims to incorporates complex scene information in the form of layouts to achieve more precise scene control. ControlNet~\cite{zhang2023adding} injects multiple modal control conditions into the model as images, enabling precise control over object structure in image generation. The control modalities in ControlNet encompass Canny edge maps~\cite{canny1986computational}, semantic maps, scribble maps, human poses~\cite{openpose}, and depth maps~\cite{birkl2023midas}. In contrast, LayerDiff does not employ complex control signal but rather relies on simple text-based control. Moreover, it introduces layer-specific prompts to achieve precise control over layer content generation. After pre-training, our LayerDiff can directly facilitate multi-level control applications.

%% file: sec/3_method.tex
\section{Preliminary on Diffusion Models}
Given an image $x_0$, diffusion models first produce a series of noisy images $x_1,...,x_T$ by adding Gaussian noise to $x_0$ according to some noise schedule given by $\bar{\alpha}_t$ as follows:
\begin{align}
\label{eq:q_sample}
    x_t = \sqrt{\bar{\alpha}_t}x_{0} + \sqrt{1-\bar{\alpha}_t} \epsilon,
\end{align}
where $\epsilon \sim \mathcal{N}\left(0, I\right)$.
Diffusion models then learn
a denoising model $\epsilon_\theta(x_t, t)$ to predict the added noise of a noisy image $x_t$ 
with the following training objective:
\begin{equation}
    \mathcal{L} = \mathbb{E}_{x_0 \sim q\left(x_0\right), \epsilon \sim \mathcal{N}(0, I), t \sim[1,T]}\left\|\epsilon-\epsilon_\theta\left(x_t, t\right)\right\|^2,
\end{equation}
where $t$ is uniformly sampled from $\{1,...,T\}$. Once the denoising model $\epsilon_\theta(x_t, t)$ is learned, 
starting from a random noise $x_T \sim \mathcal{N}(0, I)$, one can iteratively predict and reduce the noise in $x_t$ to get a real image $x_0$.
During the sampling process, we can predict the clean data $x_0$ from $\epsilon_\theta\left(x_t, t\right)$ with one-step sampling as follows:
\begin{equation}
\label{eq:epsilon_to_x0}
    \hat{x}_{0,t} = \frac{1}{\sqrt{\bar{\alpha}_t}} (x_t - \sqrt{1 - \bar{\alpha}_t} \epsilon_{\theta} (x_t, t)).
\end{equation}

\section{Method}
\label{sec:method}
In this section, we first define the data format of the multi-layered composable image and the task formulation to achieve multi-layered composable image synthesis. Then, we introduce the network architecture of LayerDiff. The layer-collaborative attention block is proposed to learn inter-layer relationships and generate the intra-layer content guided by layer-specific prompts and the layer-specific prompt enhancer is employed to further enhance the controllability of layer-specific elements. Lastly, we outline the pipeline for constructing the training dataset.
\subsection{Task Formulation}
\label{sec:task formulation}
\subsubsection{Definition of Multi-Layered Composable Image}
Compared to one entire image, a composable image $x$ is composed of multiple layers $L=\{L_i\}_0^k$, including a background layer $L_0$ and $k$ foreground layers $\{L_i\}_1^k$. A foreground layer $L_i$ is a pair of foreground image $F_i$ and foreground mask $m_i$. There is no overlap between the foreground images and foreground masks, respectively. 
The foreground mask indicates the area of the foreground object and the foreground image depicts the specific foreground object in this area. The background layer $L_0$ also includes a background image $B$ and a background mask $m_b$. The background mask is actually obtained by taking the complement of the foreground masks. 
Those foreground pairs are finally stacked together and
overlaid onto the background image to compose an entire image. 
Therefore, an image $I$ composited by multiple layers can be defined as:
\begin{align}
    I = m_b * B +  \sum_{i=1}^k \left(m_i * F_i \right)
\end{align}
where $m_b = 1 - \sum_{i=1}^k m_i$.
\subsubsection{Text-Guided Multi-Layered Image Synthesis}
\label{sec:text-guided multi-layered image synthesis.}
Unlike previous text-to-image generation tasks, which required models to generate images guided by a single comprehensive image description, multi-layered composable image synthesis not only demands a global text description to depict the overall image content but also introduces layer-specific prompts to guide the generation of specific foreground elements in each layer. 
We use the layer-specific prompt of ``the background'' for the background layer and the object descriptions as the layer-specific prompts for the foreground layers.  

We perform the diffusion process on the composition of multiple layers by adding \textit{T} steps random Guassian noies $\epsilon$ to gradually convert the original multi-layered image $x$ into a random Gauassian distribution $x_T$. LayerDiff is trained to predict the different level of noise on all layers in the opposite direction of the diffusion process under the guidance of the global prompt and layer-specific prompts. Following the LDM~\cite{rombach2022high}, we can apply the diffusion process on the latent space of the layer images $\{z_i\}_0^k$ and layer masks $\{z'_i\}_0^k$ by utilize its latent encoders respectively. In our LayerDiff, we also handle the layer masks in the RGB space by simply repeating one channel to three channel so that the layer mask encoder $\mathcal{V}$ and decoder $\mathcal{D}$ are the same as those for the layer images. Therefore, for a composable image $x$ the diffusion-based multi-layerd composable image generation loss $\mathcal{L}$ under the guidance of the global prompt $y$ and layer-specific prompts $p=\{p_0, ..., p_k\}$ can be formulated as:
\begin{equation} \label{eq:image-generation-simple-loss}
\mathcal{L} = \mathbb{E}_{\mathcal{V}(x), \epsilon \sim \mathcal{N}(0, I), t}\left[ \| \epsilon_z - \Phi_{u}(x, t, \mathcal{T}(y), \{\mathcal{T}(p)\}_0^k) \|^2 \right]
\end{equation}
where $\mathcal{T}$ is the text encoder for the global prompt and layer-specific prompts. $x$ should be a collection of layers.

In the sampling stage, LayerDiff generates all layer image latents and layer mask latents by gradually removing the noise from a random Gaussian noise signal over a finite number of steps. We can apply the latent decoder $\mathcal{D}$ to decode back the layer images and layer masks into the original data space. To obtain the predicted binary mask, we simply average values across the RGB channels for each pixel, resulting in a single-channel representation. The layer mask with maximum value is set to 1 while all other layer masks are assigned a value of 0. 
\begin{figure*}[t!]
\begin{center}
\includegraphics[width=1\textwidth]{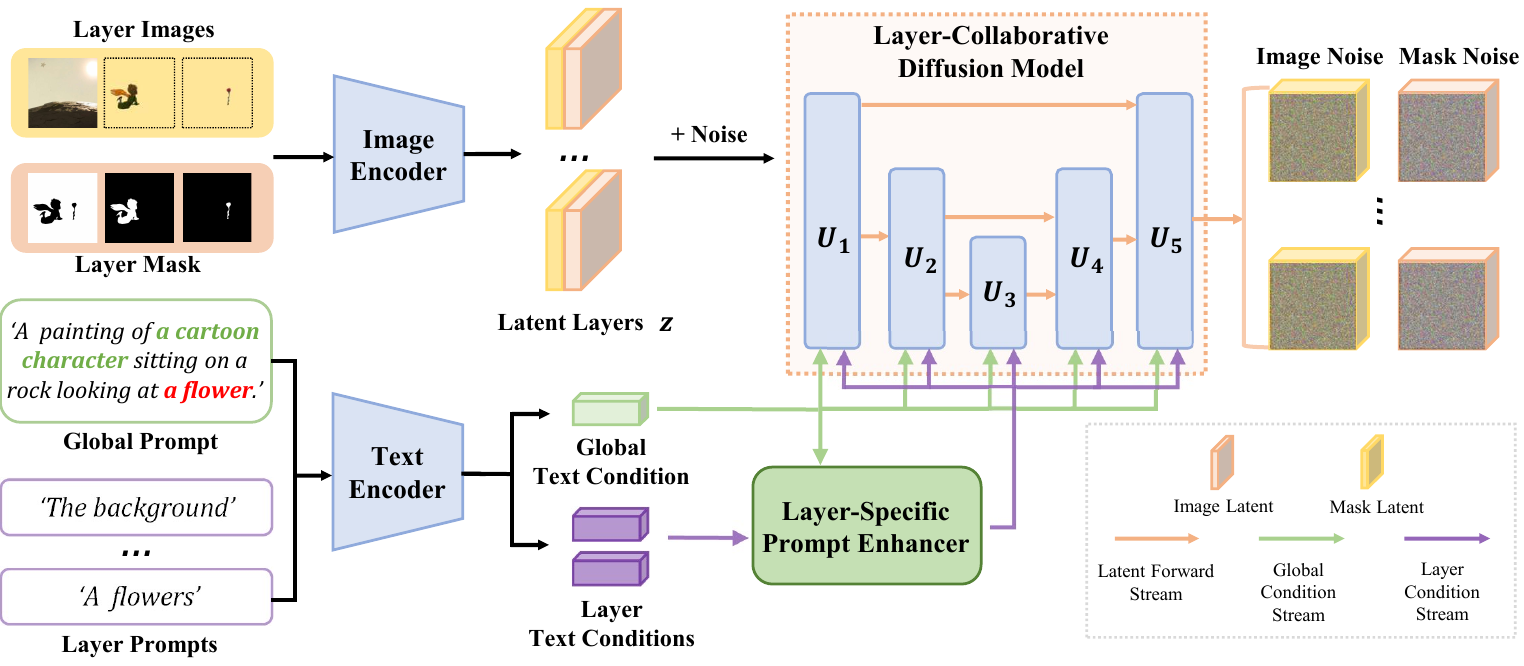}
\caption{Overall architecture of the proposed LayerDiff. LayerDiff performs the multi-layered composable image synthesis by generating the layer images and layer masks simultaneously under the guidance of both the global prompt and layer prompts. The layer-specific prompt enhancer ensures the layer text conditions to guide the content generation in each layer. In the layer collaborative diffusion model, the layer-collaborative attention block learns the cross-layer relationship and injects the text guidance signal into the model.
}
\label{fig:framework}
\end{center}
\end{figure*}
\subsection{Layer-Collaborative Diffusion Model}
\subsubsection{Architecture Overview}
As shown in Fig. \ref{fig:framework}, Our LayerDiff is designed to include the image encoder, text encoder, layer-specific prompt enhancer and a layer-collaborative diffusion model. The image encoder $\mathcal{V}$ is employed to convert the layer images and layer masks from the RGB space into the latent space. Note that we treat the layer mask as the RGB image by repeating one channel into three channel. A text encoder $\mathcal{T}$ is applied for the global prompt $c$ and layer-specific prompts $p$ to obtain the global text condition $\mathcal{T}(y)$ and layer text conditions $\{\mathcal{T}(p_i)\}$, respectively. Note that each prompt is encoded separately by the text encoder. The layer-specific prompt enhancer is proposed to ensure the information completion and controllability of each layer prompt. 
Both the global text condition and the enhanced layer text conditions are utilized to guide the generation of multi-layered composable images in the layer-collaborative diffusion model, controlling the overall content generation as well as the content generation for individual layers.
The layer-collaborative diffusion model draws inspiration from the network design in Stable Diffusion~\cite{rombach2022high}. Following the original attention block performing the guidance of the global prompt, our LayerDiff introduces the layer-collaborative attention block to learn inter-layer connections and guide the generation of content for individual layers. For the input of the layer-collaborative diffusion model, we concatenate the latent image and latent mask along the channel dimension and stack all layer latents in the layer dimension.

\subsubsection{Layer-Collaborative Attention Block}
\label{sec:layer-collaborative attention block}
Layer-collaborative attention block serves as a pivotal component in the multi-layered composable image synthesis, orchestrating the intricate interplay between different layers and guiding the generation of layer-specific content. As shown in Fig.~\ref{fig:unet_block},  the layer-collaborative attention block is structurally composed of the inter-layer attention module, text-guided intra-layer attention module and a feed-forward network (FFN). The inter-layer attention module is dedicated to learning across different layers. It processes each pixel value of the layer hidden states, capturing the relationships and dependencies between layers, ensuring that the synthesized image maintains coherence and harmony across its depth. The text-guided intra-layer attention module takes the helm when it comes to layer-specific content generation. Guided by the layer text conditions, it ensures that each layer of the image is generated in alignment with the specific textual descriptions, allowing for precise and contextually relevant layered image synthesis. The FFN further process and refine the outputs from the attention modules.
\subsubsection{Layer-Specific Prompt Enhancer}
\label{sec:layer-specific prompt enhancer}
\begin{figure}[t!]
\begin{center}
\includegraphics[width=1\linewidth]{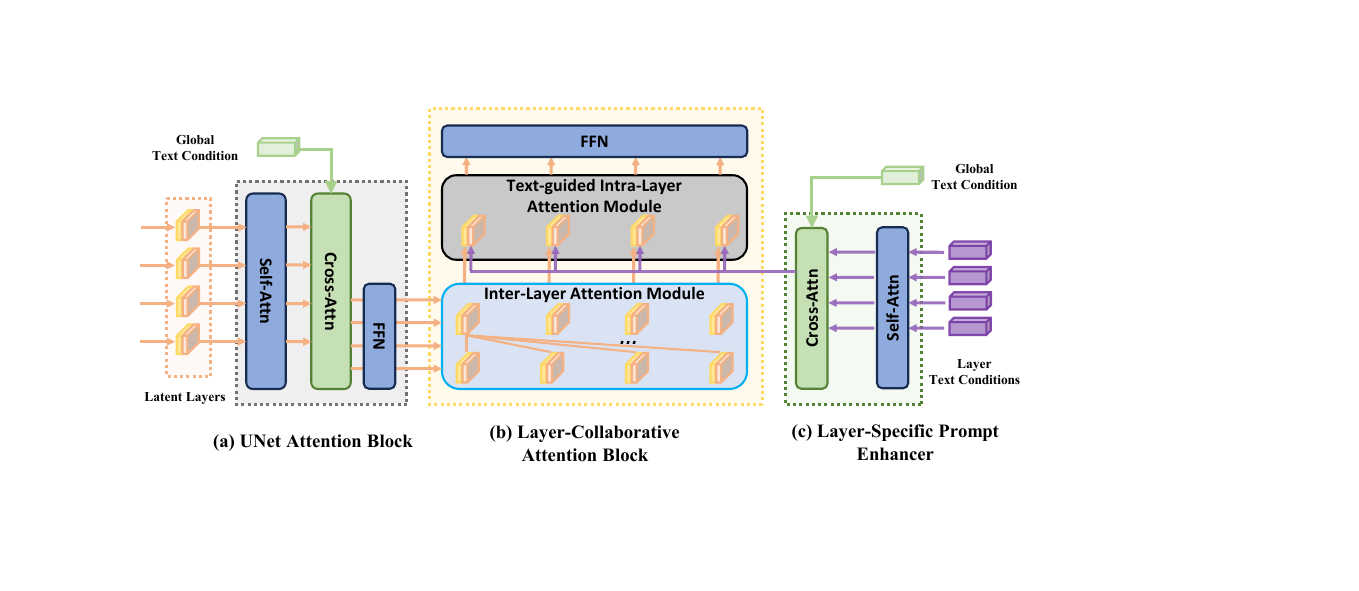}
\caption{
Detailed structure of the proposed layer-collaborative attention block. (a) The UNet attention block is commonly used in the traditional UNet Block for condition-based whole-image generation. (b) To better capture layer-wise features, we introduce the Layer-Collaborative Attention Block, which incorporates a Text-guided Intra-Layer Attention Module to guide layer content generation and an Inter-Layer Attention Module to enable cross-layer interaction. (c) The Layer-Specific Prompt Enhancer is designed to more effectively allow layer-specific prompts to assimilate information from the global prompt.
}
\label{fig:unet_block}
\end{center}
\end{figure}

Layer-specific prompt enhancer is a module designed to refine and augment layer-specific prompts by extracting and integrating relevant information from the global prompt. It potentially ensure more accurate and detailed guidance for individual layer generation in a multi-layered synthesis process. As shown in Fig.~\ref{fig:unet_block}, a self-attention layer is firstly applied to layer-specific prompts to enhance their distinctiveness from each other, further ensuring the independence of content between layers. Additionally, a cross-attention layer simultaneously takes layer-specific prompts as queries, global prompts as keys and values, aiming to enable layer-specific prompts to capture richer contextual information from the global prompt and inter-layer object relationships.
\subsubsection{Self-Mask Guidance Sampling}
\label{sec:self-mask guidance}
In the sampling stage of generating multi-layered composable images, the model may struggle to generate layer-specific content. To improve the effects of multi-layered image generation, inspired by Self-Attention Guidance Sampling~\cite{hong2023improving}, we proposed the Self-Mask Guidance (SMG) Sampling to better promote the model's focus on generating content within each layer by utilizing the predicted layer mask during the sampling.

For brevity, we illustrate the self-mask guidance sampling mechanism at $i$-th layer. 
Our goal is to guide the model to focus on the area marked by the layer mask $m_i$ of the layer image ${z_i}$. To this end, at each sampling step $t$, we conceal the information within the marked area and keep the information outside the marked area of ${z_i}_t$ as $\hat{z_i}_t$ and push the predicted noise $\epsilon_t$ for ${z_i}_t$ away from the predicted noise $\hat{\epsilon}_t$ for $\hat{z_i}_t$ as: $\tilde\epsilon_t = \hat{\epsilon}_t + (1+s)(\epsilon_t - \hat{\epsilon}_t)$, where $s$ is the guidance scale. 
$\hat{z_i}_t$ is obtained as follows:
\begin{align}
    \hat{z_i}_{0,t} &= ({z_i}_t - \sqrt{1 - \bar{\alpha}_t}\epsilon_t) / \sqrt{\bar{\alpha}_t}, \\
    \hat{z_i}_{0,t} &= \text{Gaussian-Blur}(\hat{z_i}_{0,t}), \\
    \tilde{z_i}_{0,t} &= \sqrt{\bar{\alpha}_t}\hat{z_i}_{0,t} + \sqrt{1-\bar{\alpha}_t} \epsilon, \\
    \hat{z_i}_t &= m_i \odot {z_i}_t + (1 - m_i) \odot \tilde {z_i}_t,
\end{align}
where the clean data $\hat{z_i}_{0,t}$ is first predicted with Eq. \ref{eq:epsilon_to_x0}, then apply Gaussian-Blur and $t$-step noising with Eq. \ref{eq:q_sample} to conceal the information within the marked area, and finally recover the information outside the marked area by merging with ${z_i}_t$ according to the predicted layer mask $\hat{m_i}$ at each step. The predicted layer mask $m_i$ is derived by predicting $z_0'$ using Eq. \ref{eq:epsilon_to_x0} and decoding back to the RGB space and convertig it into binary mask as mentioned in Sec. \ref{sec:text-guided multi-layered image synthesis.}.

\begin{figure*}[t!]
\begin{center}
\includegraphics[width=\textwidth]{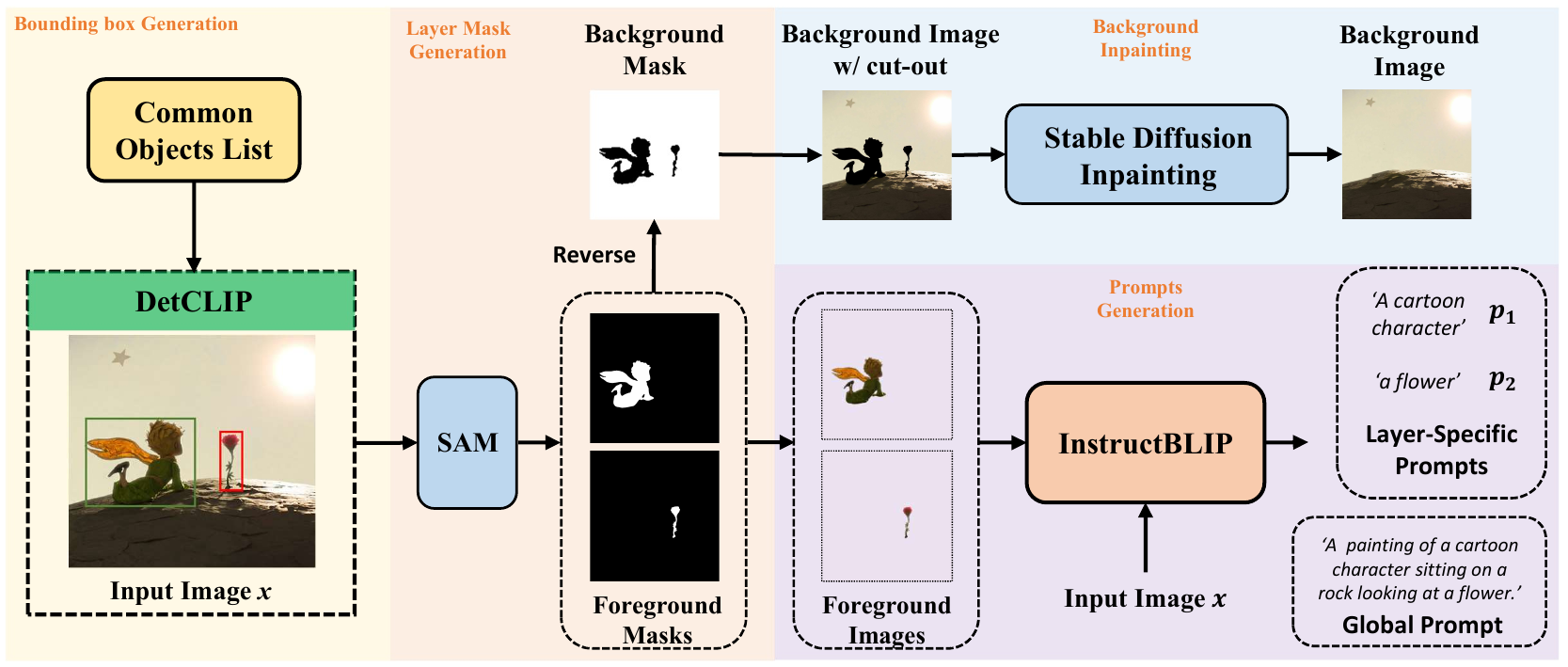}
\caption{Pipeline of the Multi-Layered Composable Image Construction. We use the InstructBLIP~\cite{dai2023instructblip} for image captioning. These prompts guide open-set segmentation via DetCLIP+SAM to produce image layers and masks and the background image is refined by using the Stable Diffusion inpainting model~\cite{rombach2022high}.}
\label{fig:data_construct}
\end{center}
\vspace{-6mm}
\end{figure*}
\subsection{Dataset Construction}
\label{sec:dataset-construction}
In Fig. \ref{fig:data_construct}, we present the pipeline we used to construct multi-layered composable images. To construct accurate foreground layers, we collect a common object list selected from the Object365's~\cite{shao2019objects365} categories and employ a powerful detector, DetCLIP~\cite{detclip}, to precisely locate all foreground objects in the image. After obtaining instance-level bounding boxes of the foreground objects, we use SAM~\cite{kirillov2023segment} to obtain the foreground object masks within the corresponding boxes, resulting in precise foreground object layers. By inverting the foreground masks, we obtain the background layer mask and the corresponding cut-out background image. We use the stable diffusion inpainting model~\cite{rombach2022high} to fill in the hollowed-out area to obtain the complete background image. To enable the model to better learn the alignment between layers and text, we employ InstructBLIP~\cite{dai2023instructblip} to generate the precise and contextually-rich description of the whole image as the global prompt and the short but accurate object description for each foreground layer as the layer prompts. Note that for the background image, we simply use ``the background'' as the layer prompt. Finally, we produce the training sample consisting of global prompt, layer images with their corresponding layer prompts and masks. Details of the pipeline can be found in Appendix.

    
By employing the well-designed proposed data construction pipeline, we obtain the high-quality, multi-layered, composable image dataset, named MLCID. We collect the training set including 1M data from the LAION400M dataset~\cite{schuhmann2021laion} and an additional 1M data from the private dataset. We only consider the composable image with two, three, and four layers. Furthermore, we construct the testing set with 27k images and prompts by adopting the proposed data construction pipeline, ensuring it remains entirely distinct from our training set. The quantities of data for two, three, four layers are 1.7M, 0.3M and 0.08M, respectively.

%% file: sec/4_experiment.tex
\section{Experiments}
\label{sec:experiments}
In this section, we first describe the implementation details of our LayerDiff (Sec. \ref{sec:implementation details}) and experimental details (Sec.~\ref{sec:experimental details}). In our quantitative results (Sec.~\ref{sec:quantitative results}), we compare the Stable Diffusion of the text-to-image synthesis on private benchmark and public benchmark and conduct the ablation study of proposed components. In the quantitative results (Sec. \ref{sec:qualitative results}), we visualize the quality of the multi-layered composable image synthesis and explore LayerDiff's application, including layer inpainting and layer style transfer.
\subsection{Implementation Details}
\label{sec:implementation details}
To expedite the convergence of our model, we initialized our autoencoder and 3D U-Net using the weights from Stable Diffusion v1.5~\cite{rombach2022high}. The latent mask passes through a Convolutional layer and follows a zero-initialized scalar parameter, controlling the information injection to the image branch by directly adding the mask information on the image per pixel. More detailed parameter initialization rules can be found in the Appendix.
During the training process, we apply the same timestep $t$ to all layer with 50\% probability and a 50\% probability of applying different timesteps $t$ for different layers. To prevent a decline in generation performance due to the overly tight segmentation of foreground objects by the masks, we employ a dilation operation on the masks when obtaining cutting-out foreground images. The dilation kernel size is $5\times5$. However, during training, the input masks are used without the dilation operation.
\begin{table*}[t!]
\center
\setlength{\tabcolsep}{5pt}
\renewcommand\arraystretch{0.92}
\caption{Main results on text-to-image synthesis. Our LayerDiff can perform multi-layered composable image synthesis. LayerDiff can achieve text-to-image synthesis by composing the multiple layers into one composite image. ``CS'' indicates the CLIP-Score~\cite{hessel2021clipscore}.}
\vspace{-6mm}
\resizebox{\textwidth}{!}{
\begin{tabular}{l|ccc ccc ccc}
\toprule
  &  \multicolumn{3}{c|}{Two Layers} &  \multicolumn{3}{c|}{Three Layers} & \multicolumn{3}{c}{Four Layers}   \\
    
  &  FID   & Layer CS & \multicolumn{1}{c|}{CS}  &  FID   & Layer CS & \multicolumn{1}{c|}{CS} &  FID   &  Layer CS & \multicolumn{1}{c}{CS}  \\
\midrule
Stable Diffusion V1.5 & 27.0 &  -  &  33.5   &  25.9  & -   & 33.9   & 32.8  & - & 33.8   \\
LayerDiff       & 35.1 & 31.2  &  33.7   & 58.9  & 25.9   & 32.2   & 83.9 & 23.7  & 30.5   \\
LayerDiff+SMG   & 21.3 & 33.1  &  35.5   & 42.8   & 27.3   & 33.7   & 58.7  & 25.4 & 32.3   \\

\bottomrule         
\end{tabular}
}
\vspace{-4mm}
\label{tab:main-result-table}
\end{table*}

\begin{table*}[t]
\center
\setlength{\tabcolsep}{5pt}
\Large
\caption{Ablation study on the proposed components. All evaluations are performed on the test set from LAION-400M.  $\mathcal{S}$ means the self-attention layer. $\mathcal{C}$ is the cross-attention layer. ``CS'' indicates the CLIP-Score~\cite{hessel2021clipscore}. Layer CLIP-Score evaluates the alignment between the layer prompts and foreground layers. 
``Mask Dilate'' means the dilation kernel size of the mask. ``Mask Info Injection'' means the module to control the mask information injected into the UNet.}
\vspace{-8mm}
\resizebox{\textwidth}{!}{
\begin{tabular}{lccc|ccc|ccc|ccc}
\toprule
 Layer-Specific & Intra-Layer &  Mask Dilate & Mask Info &  \multicolumn{3}{c|}{Two Layers}    & \multicolumn{3}{c|}{Three Layers} &   \multicolumn{3}{c}{Four Layers} \\
 Prompt Enhancer   & Attention  &  & Injection  &  FID  & Layer CS & CS & FID  & Layer CS & CS & FID & Layer CS & CS  \\
\midrule
None                           & \xmark      & 5   & FC   &  358.0 &  17.5  &  17.0  & 221.6 &  18.7 &  19.9 &  238.5 & 18.6 &  19.3    \\
None                           & \cmark      & 0   &  FC   &  53.7  &  29.2  &  30.9   &  102.3  & 23.2 & 27.1   &  144.7  &  21.5 & 24.4  \\
None                           & \cmark      & 5   &  FC   &  45.8 &  29.6  &  31.2   &  92.2  & 24.5 &  29.0  & 142.2 & 22.1 & 26.4 \\
$\mathcal{S}$                   & \cmark     & 5   &  FC   & \textbf{44.4}  & 29.8  & 31.5  &  94.4   &  24.6  & 29.1 &  143.2  &  22.4  & 26.7  \\
$\mathcal{S}$ + $\mathcal{C}$    & \cmark  & 5   &  FC   &  46.7   & \textbf{ 29.9 } & \textbf{31.6} & 92.9 & 24.9  & 29.4 &  140.2  &  22.4  &  26.9 \\
$\mathcal{S}$ + $\mathcal{C}$   & \cmark  & 5   &  Scalar   &  50.1   &  29.8  &  31.5   &  \textbf{79.8}  & \textbf{25.6} &  \textbf{30.3}  &  \textbf{112.8}  & \textbf{23.1} & \textbf{28.3} \\
\bottomrule         
\end{tabular}
}
\vspace{-8mm}
\label{tab:ablation}
\end{table*}

\subsection{Experimental Details}
\label{sec:experimental details}
\vspace{-2mm}
We pre-train models on the MLCID with about 2M multi-layered image-text pairs. The resolution of the image is $256 \times 256$. Following Stable Diffusion, the pixel values of the layer images and layer masks are all normalized to [-1, 1] and the autoencoder and text encoder are frozen. We apply the AdamW~\cite{loshchilov2017decoupled} optimizer with a weight decay of 1e-2. The learning
rate is set as 1e-5 with 500 steps linear warmup and kept
unchanged until the training is finished. The batch size is set as 32 and we apply 8 steps of gradient accumulation to stabilize the training and obtain better generalization performance. We randomly drop 10\%
global text condition and layer text condition in our pre-training. 

In our sampling stage, we apply DDIM Scheduler~\cite{song2020denoising} with 50 iterations and both classifier-free guidance~\cite{ho2022classifier-free} scale and the proposed SMG scale of 3. Specifically, instead of using the empty text as the unconditional text, we use the concatenation of the foreground layer texts as the negative global text condition and the negative layer text condition for the background layer. Similarly, we use the background prompt, i.e., ``the background'' as the negative text condition for the foreground layer. 

Evaluation: We evaluate our LayerDiff on the MLCID's testing dataset with 27k samples. 
We use the open-source tool, torch-fidelity~\cite{obukhov2020torchfidelity}, to evaluate the metrics of FID. We also apply the CLIP-score to evaluate the alignment between generated composable images and the given prompts by using the RN50 backbone of CLIP~\cite{radford2021learning}. We also calculate the layer CLIP-Score by calculating the similarity between the foreground image and the corresponding layer prompt. We compare to Stable Diffusion~\cite{rombach2022high} by directing generating images of a resolution of 512 and then resizing them to 256.

\vspace{-2mm}
\subsection{Quantitative Results}
\label{sec:quantitative results}
\noindent \textbf{Main Results.} Table~\ref{tab:main-result-table} demonstrates our main results on the MLCID. Our LayerDiff's performance on two-layered image synthesis is comparable to Stable Diffusion's whole-image synthesis performance.
In tests involving a greater number of layers and the full-test set of MLCID, we observed a performance disparity in LayerDiff.
We speculate that this gap stems from the insufficient training of our model on three and four layers. Within our MLCID, there are merely around 15\% and 3\%, respectively. This inadequacy in the data volume for three-layer and four-layer images hinders the model's performance in generating images with these specific layer counts. In the future, we aim to enhance the generalization ability of our model by enlarging the scale of the training dataset.

\begin{figure*}[t]
\begin{center}
\includegraphics[width=\textwidth]{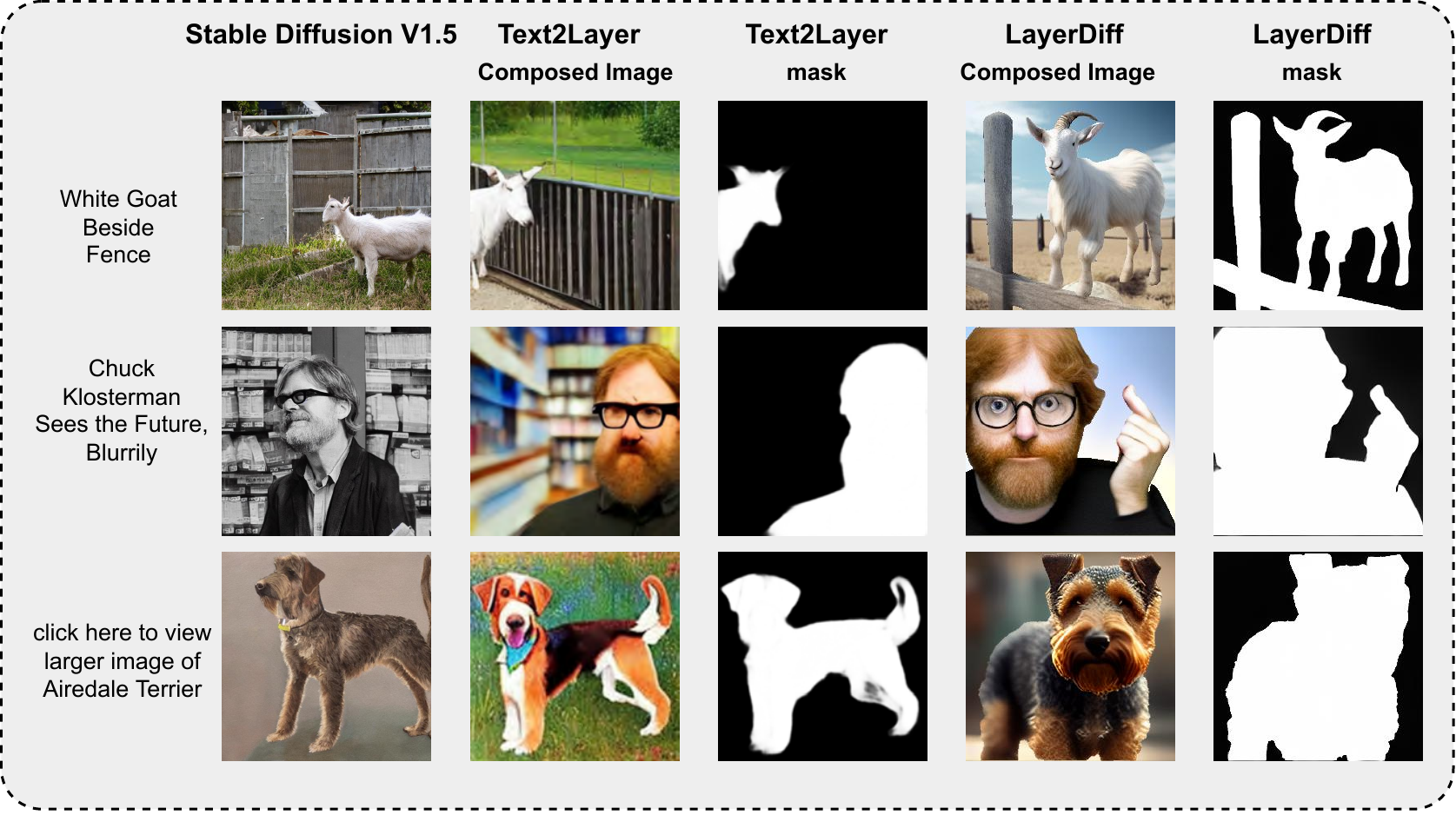}
\vspace{-5mm}
\caption{
Qualitative results of the synthesized two-layered images. We compare our composited image with the samples generated by Stable Diffusion~\cite{rombach2022high} using the whole-image generation approach, and the Text2Layer~\cite{zhang2023text2layer} that also performs the multi-layered image generation. 
The quality of our multi-layer generation is found to be comparable to the samples produced by Stable Diffusion.
}
\label{fig:visualize-for-multi-layered-image}
\end{center}
\vspace{-10mm}
\end{figure*}

\noindent \textbf{Ablation Study.}~We conduct the ablation study on a subset with 0.2M data of the MLCID and all models are pre-trained about 50k iterations. 
From the result of ablation study on Table~\ref{tab:ablation}, it is discerned that the layer-specific prompt plays a crucial role in multi-layer generate representation of three-layered images and four-layered images. The layer-specific prompts assists in guiding the model to generate content for each layer and establish inter-layer relationships. Furthermore, the layer-specific prompt enhancer, utilizing the self-attention and cross-attention layers will help the model capture the pattern on three- or four-layer generation. Additionally, employing dilated layer masks to obtain layer images proves to be beneficial for multi-layer generation. Moreover, Table~\ref{tab:main-result-table} demonstrates that SMG can unleash the power of the model and greatly enhance the generation quality.

\begin{figure*}[t]
\begin{center}
\includegraphics[width=\textwidth]{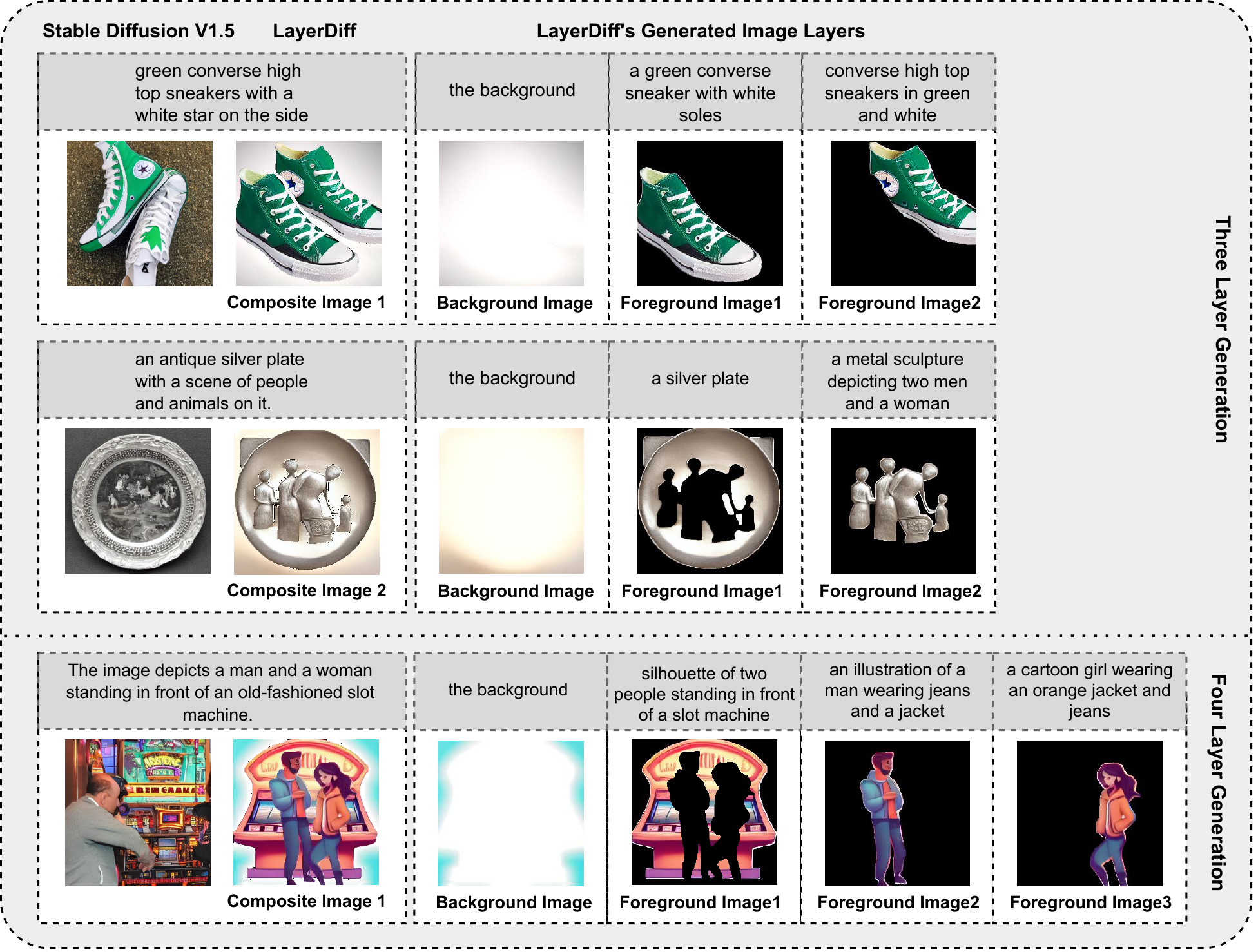}
\vspace{-4mm}
\caption{
Qualitative results of the synthesized three-layered and four-layered images.
}
\label{fig:visualize-for-multi-layered-image-three-four}
\end{center}
\vspace{-8mm}
\end{figure*}

\begin{figure}[t]
\vspace{-2mm}
\begin{center}
\includegraphics[width=\linewidth]{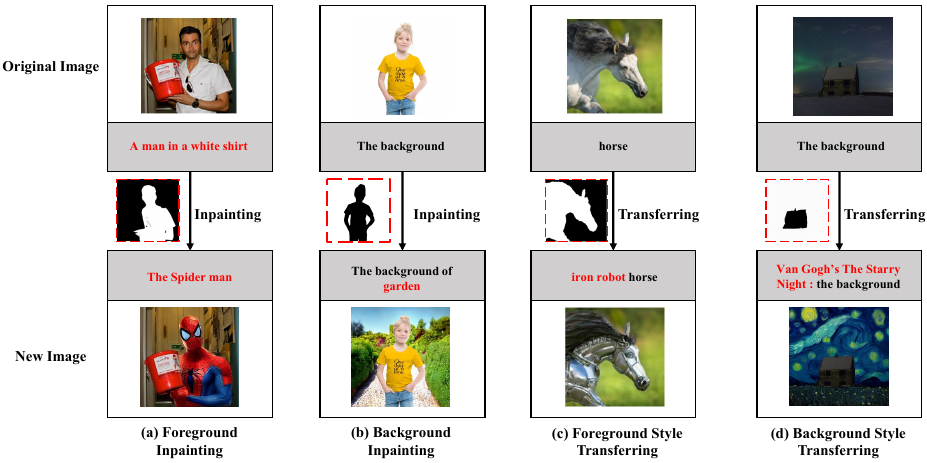}
\vspace{-4mm}
\caption{Four applications of LayerDiff. (a) and (b) demonstrate the foreground layer inpainting and background layer inpainting, respectively. The layer inpainting regenerates the target layer guided by layer prompts to obtain desired content. (c) and (d) showcase foreground layer style transferring and background layer style transferring, respectively. Layer style transferring transforms the target layer to a new style by adding additional control signal.
The red text denotes modified text or additional control signals, affecting both global and target layer prompts. 
}
\vspace{-8mm}
\label{fig:visualize-layer-inpainting-styleTransferring}
\end{center}
\end{figure}

\vspace{-3mm}
\subsection{Qualitative Results}
\label{sec:qualitative results}
Figure~\ref{fig:visualize-for-multi-layered-image} and Figure~\ref{fig:visualize-for-multi-layered-image-three-four} displays the layer generation in our model and visualizes the produced individual layer images, layer masks, along with the final results in comparison with those images generated by Stable Diffusion~\cite{rombach2022high} and Text2Layer~\cite{zhang2023text2layer}. Through visualization, it is evident that our LayerDiff demonstrates the capability to generate multi-layered composable images. By overlaying and composing the layer images and image masks, the composite images generated by LayerDiff achieve competitive results when compared to the whole-image generation method, Stable Diffusion.

\vspace{-2mm}
\subsection{Application}
\label{sec:application}

After pre-training, our model naturally supports various applications without any fine-tuning, including layer inpainting and layer style transferring. 

\noindent \textbf{Layer Inpainting.} By fixing certain layers and modifying the layer prompt of a specific layer, we can facilitate the regeneration of that layer from noise. Figure \ref{fig:visualize-layer-inpainting-styleTransferring} illustrates the effects of foreground layer inpainting and background layer inpainting using LayerDiff. 
LayerDiff can specify any layer to accurately regenerate specific content on that layer, achieving precise layer inpainting. Concurrently, by allowing the non-target layers' mask changing, LayerDiff generates natural content on the target layer, e.g., the shape of the sleeves of the man in the foreground inpainting will not affect the shape of Spider-Man's arm.

\noindent \textbf{Layer Style Transferring.} We can achieve layer-level style transfer. By introducing a certain degree of noise to a specific layer and appending stylized text to the original layer prompt, we instruct the model to perform denoising sampling for the specified layer, thereby accomplishing style transfer for that layer. For instance, as depicted in Fig. \ref{fig:visualize-layer-inpainting-styleTransferring}, we can induce a Van Gogh style transformation for the background by specifying the style through the layer prompt.


%% file: sec/5_conclusion.tex
\vspace{-3mm}
\section{Conclusions}
\vspace{-2mm}
In conclusion, we introduce LayerDiff, a model designed for text-guided, multi-layered image synthesis. Unlike traditional whole-image generative models, LayerDiff enables layer-wise generation by leveraging layer-collaborative attention modules. Empirical results show that LayerDiff achieves comparable image quality while offering greater flexibility and control. This work not only extends the capability of text-driven generative models but also paves the way for future advancements in controllable generative applications, e.g., layer-specific editing and style transfer.

\noindent \textbf{Limitation.} Existing multi-layer training data generation pipelines are inefficient, hindering the capability to produce large-scale training data effectively. This inefficiency compromises the performance of the resulting models, preventing them from achieving impressive results. Therefore, optimizing the design of efficient multi-layer training data pipeline represent promising future research directions.

%% file: sec/X_suppl.tex

\section{Implementation Details}
We initialize our model parameters by utilizing the pre-trained Stable Diffusion. We randomly initialize the inter-layer attention module and the layer-specific prompt enhancer. The text-guided intra-layer attention module is initialized by the original cross-attention layer in Stable Diffusion. To ensure the effectiveness of the parameters initialized from Stable Diffusion, we zero-initialize the weight of the last linear layer of each module in the layer-collaborative attention block and layer-specific prompt enhancer. Additionally, We initialize the input convolutional layer and output convolutional layer for the layer mask using the weights from the stable diffusion for images. We employ a learnable zero-initialized scalar parameter to control the mask information injection into the layer image after the input convolutional layer. 

During the Self-Mask Guidance sampling process, we set the kernel size and sigma of the Gaussian blur to 31 and 3, respectively.

\section{Data Construction Details} 

To mitigate the issue that the detector generating undesired bounding boxes for background elements, we first filter the bounding box with score lower than 0.1 and then conduct a comparison between the masks obtained from SAM and the foreground mask predicted by the saliency detector, ICON~\cite{zhuge2021salient}. Only masks with an overlap rate exceeding 10\% are retained. Additionally, we eliminate masks that are encompassed by larger ones to prevent overlap. We preserve the masks whose areas range from 1\% to 80\% of the total image size.

For the global prompts generation, we input the entire image and employ the instruction of ``Please provide a comprehensive description of the image, including the color, shapes, and any actions taking place.''. For the layer prompt generation, we use the instruction of ``Briefly describe the main subject of the image.'' and input the cropped foreground images based on the foreground masks.

During the Stable Diffusion inpainting process, we employ an empty text condition as the prompt. Concurrently, terms related to foreground objects are used as negative prompts to refine the output.

\section{Dataset Details}
\begin{figure*}[t!]
\vspace{-1em}
\begin{center}
\includegraphics[width=1\linewidth]{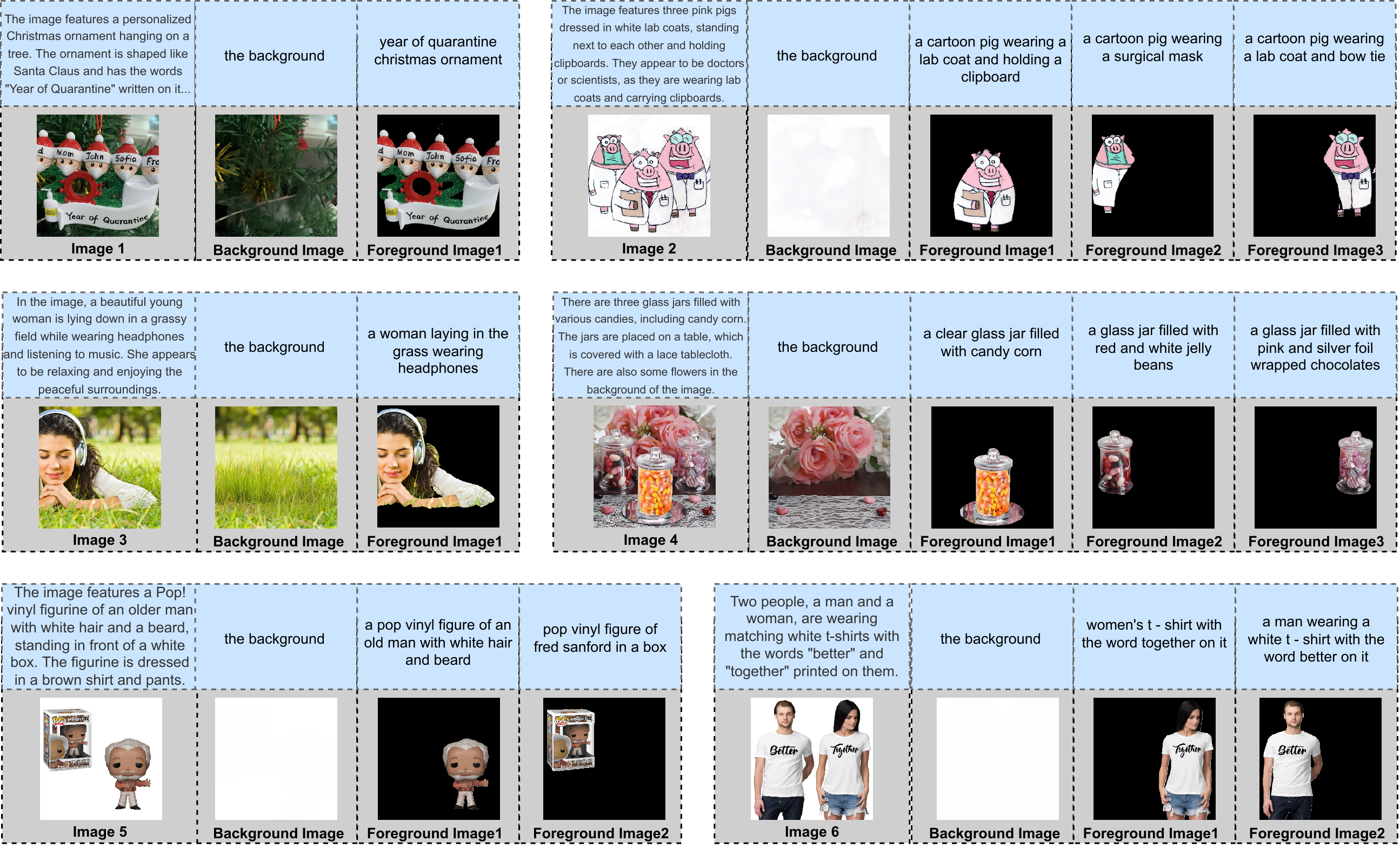}
\caption{Samples from the constructed multi-layered composable image dataset are presented. We visualize the two-layered, three-layered, and four-layered images.}
\label{fig:appendix-visualize-dataset-samples}
\end{center}
\vspace{-4mm}
\end{figure*}

Our MLCID is a long-tailed dataset comprising 1.7M two-layer images, 0.3M three-layer images, and 0.08M four-layer images. The lack of sufficiently extensive data results in the model lacking generalization capability in generating three-layer and four-layer images. In the future, we consider expanding the scale of multi-layered image by using the synthesized data. 
We visualize several samples from the our MLCID in Fig.~\ref{fig:appendix-visualize-dataset-samples}.
\begin{figure}
\begin{center}
\includegraphics[width=0.9\linewidth]{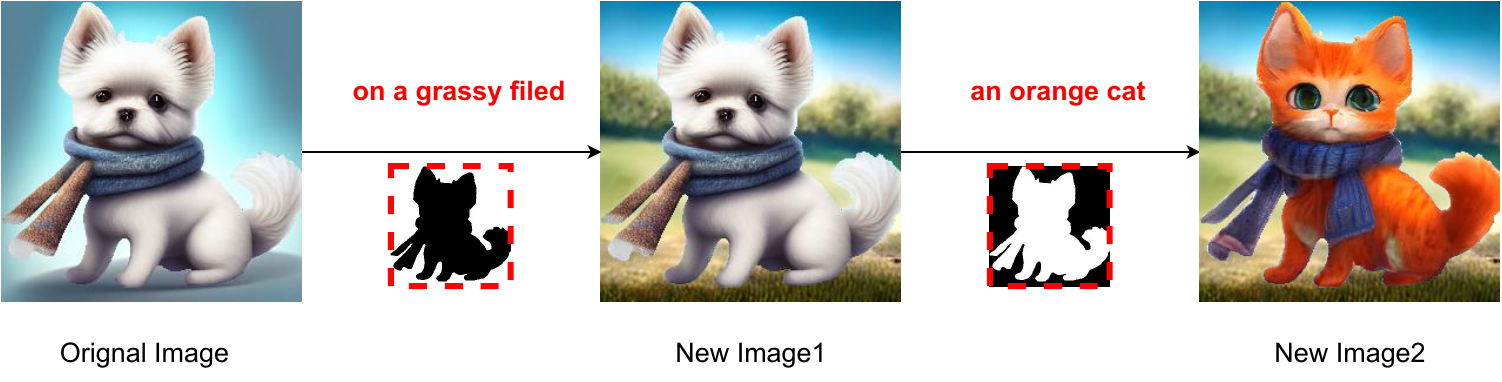}
\vspace{-2mm}
\caption{More examples of the inpainting. We first perform the background layer inpainting and then apply the foreground layer inpainting.}
\label{fig:appendix-visualize-inpainting}
\end{center}
\vspace{-8mm}
\end{figure}
\begin{figure}
\begin{center}
\includegraphics[width=1\linewidth]{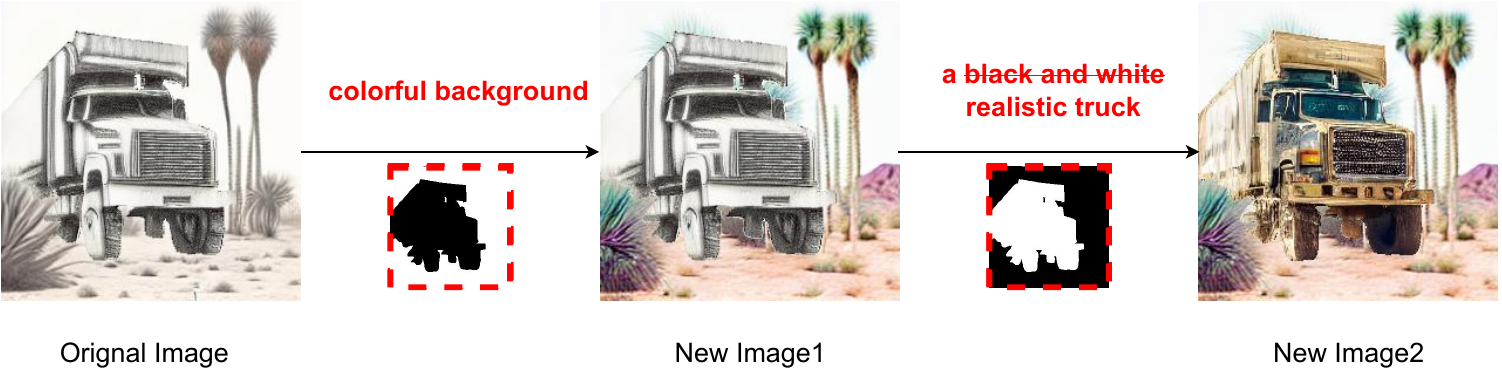}
\vspace{-5mm}
\caption{More examples of the style transfer. We first perform the background layer style transfer and then apply the foreground layer transfer.}
\label{fig:appendix-visualize-style-transfer}
\end{center}
\vspace{-8mm}
\end{figure}
\section{Application Details}
For the layer inpainting, the non-target layers are first encoded by the image encoder to transform into latent space. We add T-step noises on the non-target layers. The target layers are sampled directly from the Gaussian distribution. In each denoising step, we replace the predicted $x_0$ of non-target layers, as described in the formula (12) of \cite{song2020denoising}, with the original latent layer. Using this operation, we maintain the latent representations of the non-target layer images unchanged. The non-target layer masks' latent representations remain unchanged within \textit{k} steps but change in subsequent steps. This ensures that the masks of the non-target layers consistently adapt to the new target layers. \textit{k} is set to 1 in our experiments.


For layer style transfer, akin to the SDEdit~\cite{meng2021sdedit}, we introduce a strength of 0.8 on all layers and perform denoising 80\% of the sampling steps. During the denoising processing, we replace predicted $x_0$ with the original $x_0$ on both the layer images and layer masks of non-target layers.

\section{More Visualizations}
\begin{figure*}
\begin{center}
\includegraphics[width=0.9\textwidth]{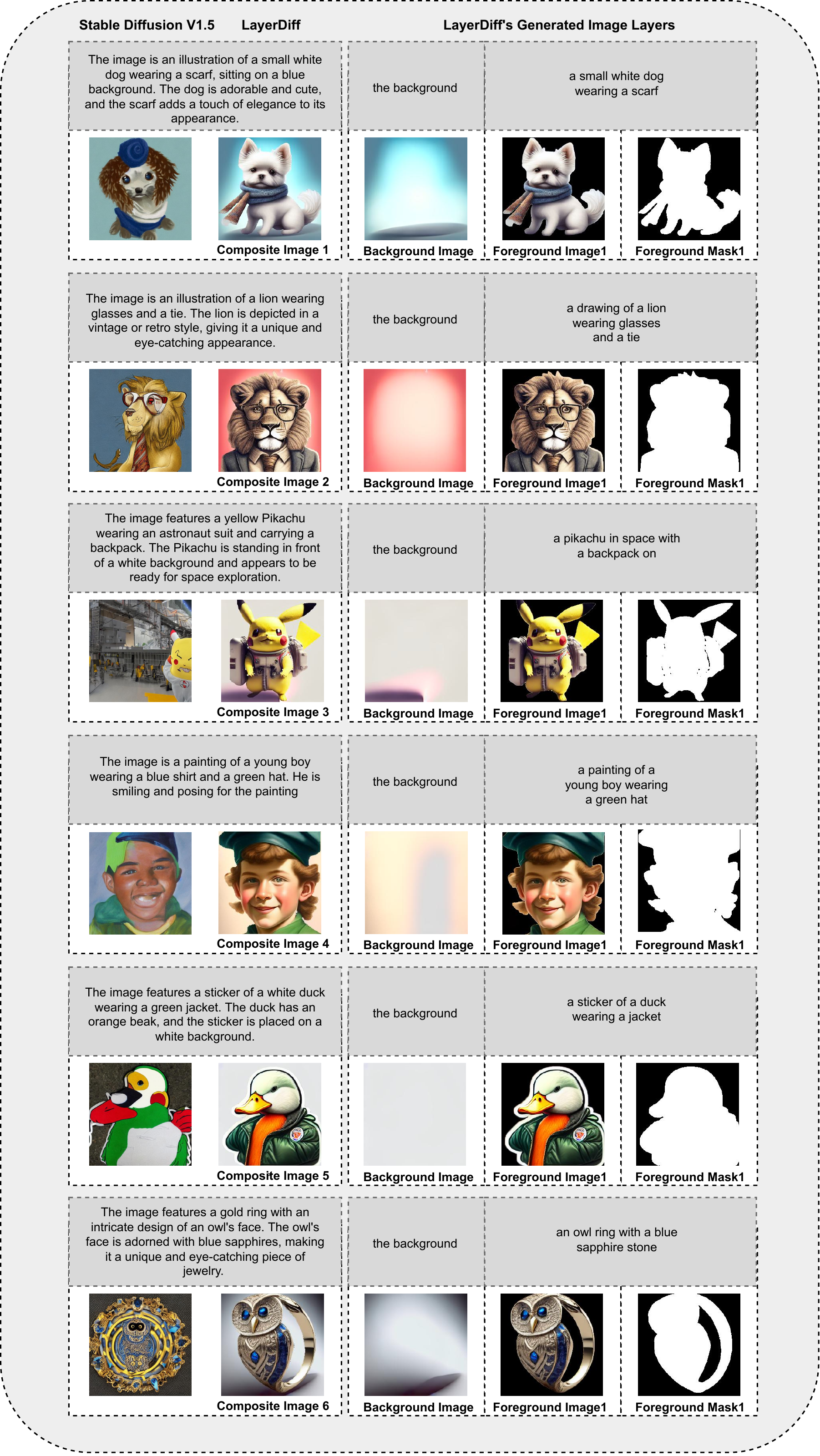}
\caption{Synthesized two-layered images.}
\label{fig:appendix-visualize-two-layer}
\end{center}
\end{figure*}

\begin{figure*}
\begin{center}
\includegraphics[width=1\textwidth]{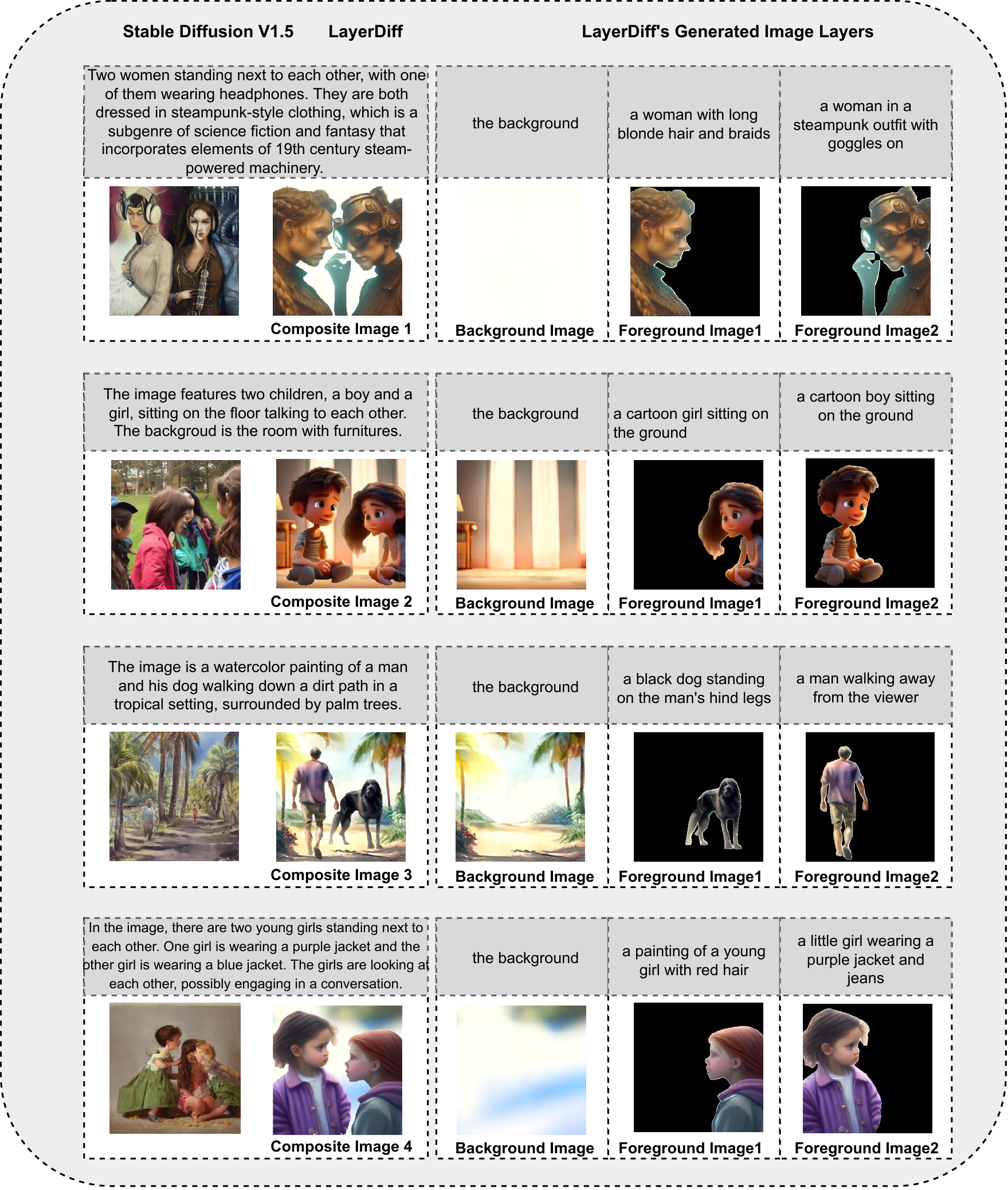}
\caption{Synthesized images with more than two layers.}
\label{fig:appendix-visualize-three-four-layer}
\end{center}
\end{figure*}

In Fig.~\ref{fig:appendix-visualize-two-layer} and Fig.~\ref{fig:appendix-visualize-three-four-layer}, we present additional examples of multi-layered composable images generated by our model. These examples are juxtaposed with samples generated using Stable Diffusion, a method oriented towards whole-image generation. This comparison aims to showcase our layered generation approach's distinct capabilities and potential advantages over traditional whole-image generation techniques. 

Furthermore, in Fig.~\ref{fig:appendix-visualize-style-transfer} and Fig.~\ref{fig:appendix-visualize-inpainting}, we showcase additional applications of LayerDiff in the contexts of style transfer and inpainting. These examples further demonstrate the versatility and efficacy of our approach in diverse image manipulation tasks.
\begin{table*}[ht!]
\vspace{-4mm}
\center
\setlength{\tabcolsep}{5pt}
\renewcommand\arraystretch{0.92}
\small
\caption{The ablation of the CFG, SAG, and SMG. ``CFG'' is the classifier-free guidance~\cite{ho2022classifier-free}. The scale of CFG, SAG, SMG is 3. The experiments are conducted on a 1k prompts set that randomly sampled from the test set of MLCID. The distribution of prompts for generating images included 500 for two-layered, 300 for three-layered, and 200 for four-layered images.}
\vspace{-4mm}
\resizebox{\textwidth}{!}{
\begin{tabular}{l|ccc|ccc|ccc|ccc}
\toprule
  &  \multicolumn{3}{c|}{Two Layers}    & \multicolumn{3}{c|}{Three Layers} &  \multicolumn{3}{c}{Four Layers} & \multicolumn{3}{|c}{All} \\
 Method & FID  & Layer CS & CS & FID  & Layer CS & CS & FID & Layer CS & CS & FID & Layer CS & CS \\
\midrule
 CFG    & 128.6 & 33.3 & 35.8 & 195.8 & 26.2 & 32.7 & 246.8 & 24.5 & 31.1 & 112.8 & 29.4 & 34.0 \\
 CFG+SAG  & 112.5 & 34.7 & 36.9  & \textbf{174.4} & 27.0 & 33.8  & 211.2 & 25.0 & 32.2 & 92.6 & 30.7 & 35.0 \\
 CFG+SMG  & \textbf{108.9} & \textbf{35.2} & \textbf{37.5} & 175.0 & \textbf{27.5} & \textbf{34.2} & \textbf{204.3} & \textbf{25.8} & \textbf{32.5} & \textbf{88.9} & \textbf{31.0} & \textbf{35.5} \\
\bottomrule         
\end{tabular}
}
\vspace{-4mm}
\label{tab:appendix-ablation-sag-smg}
\end{table*}

\begin{figure*}[h]
\vspace{-1em}
\begin{center}
\includegraphics[width=1\linewidth]{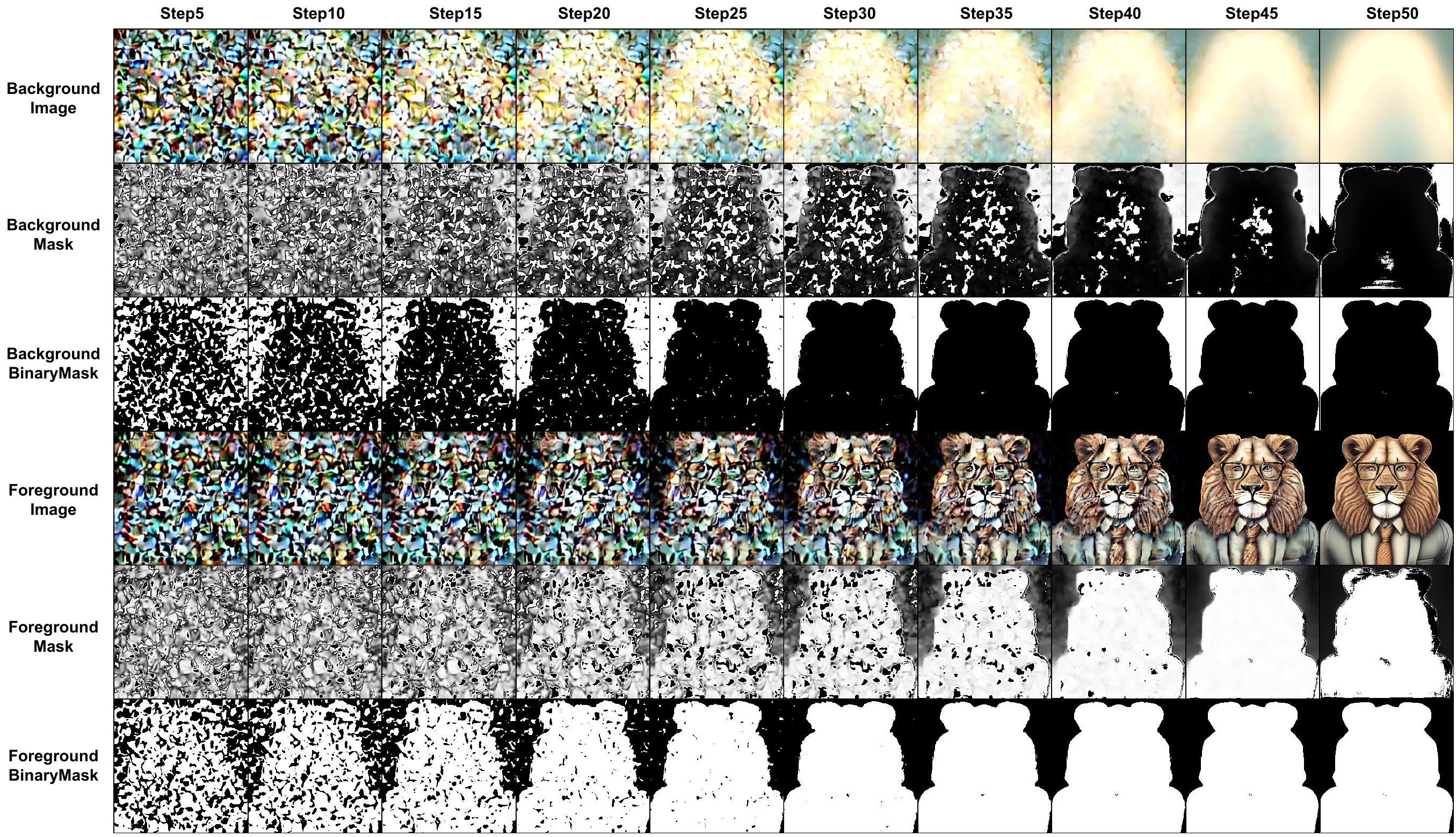}
\vspace{-2em}
\caption{Visualization of the LayerDiff's sampling process, conducted over 50 steps with intervals of visualization at every fifth step.}
\label{fig:appendix-visualize-sampleing-process}
\end{center}
\vspace{-4mm}
\end{figure*}

Figure ~\ref{fig:appendix-visualize-sampleing-process} visualizes the LayerDiff's sampling process. It is observable that the model establishes the structural relationships between layers early in the process, with a focus on refining textures in the later stages of sampling. Furthermore, the generation of the mask does not perfectly produce binary images. Accurate layer masks are obtained by comparing the values of masks between layers to binary the mask images. Besides, we visualize the attention map of the inter-layer attention and text-guided intra-layer attention in Fig. \ref{fig:appendix-visualize-attention-map}. We posit that intra-layer attention focuses on conveying textual guidance to each layer, while inter-layer attention discards irrelevant information from the current layer based on observations across layers. 

\section{More Applications}
\subsection{Sampling with Mask Priors}
\begin{figure}[ht!]
\vspace{-1em}
\begin{center}
\includegraphics[width=0.9\linewidth]{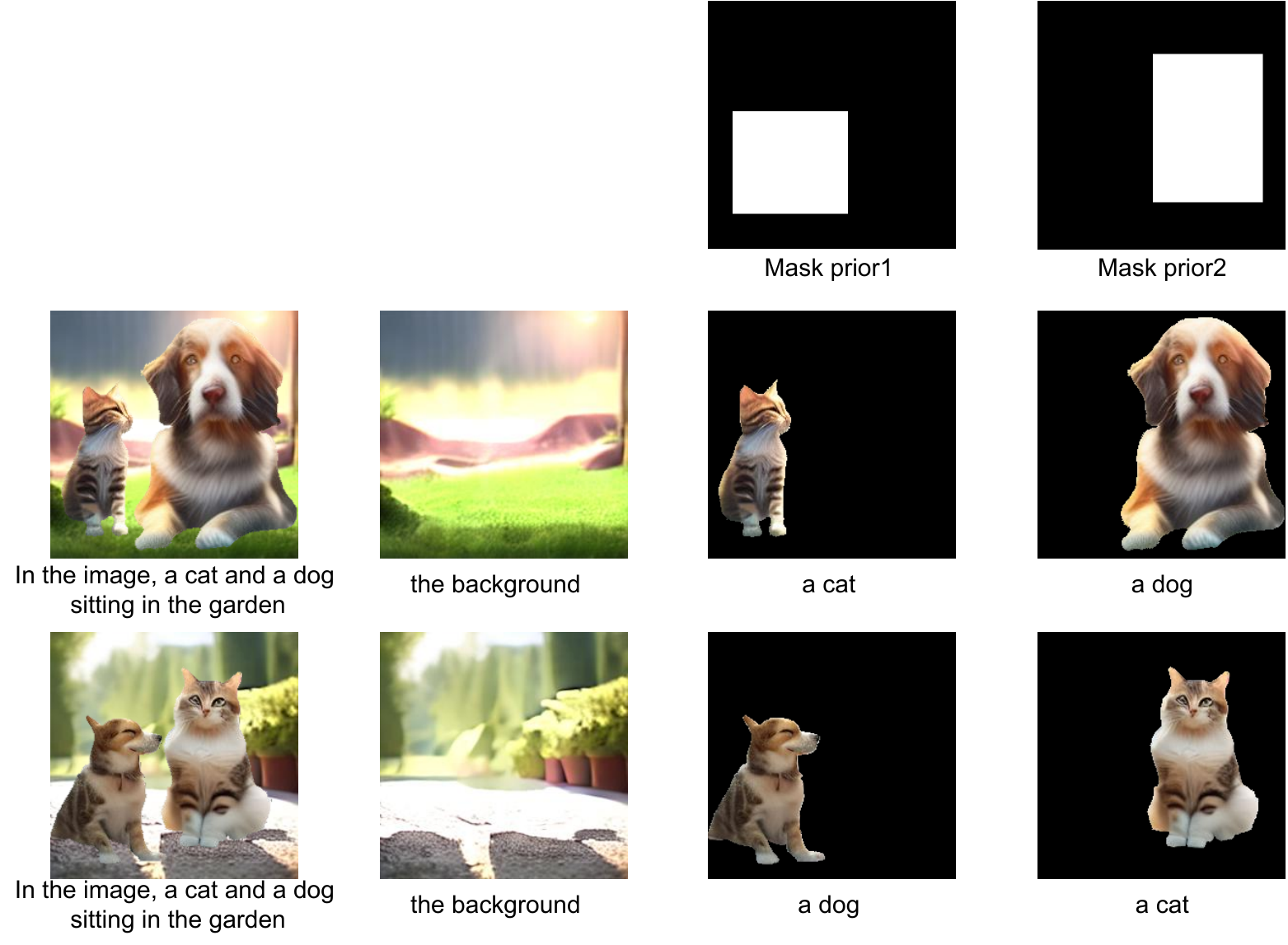}
\vspace{-1em}
\caption{Synthesized samples by sampling with mask priors. Given the boxes as the mask priors for foreground layers, we can import the coarse spatial relationship and the relative size of objects between layers. }
\label{fig:appendix-visualize-layout-control}
\end{center}
\vspace{-3mm}
\end{figure}

Sampling with mask priors indicates initializing $x_T$ with the given layer masks. This method provides the model with spatial information about the location and approximate size of foreground objects within each layer. The mask priors can be constructed using bounding boxes or object masks of foreground objects, offering priors of different granularities. Specifically, a rough bounding box is assigned to each foreground layer as the mask prior, while the mask prior for background layers is obtained by reversing the summation of layer masks of the foreground layers. Initially, the mask priors are encoded into the latent space using the image encoder and added T steps noise to serve as the initialization for the layer mask latent, while the layer image latent is sampled from a Gaussian distribution. Sampling proceeds according to the standard sampling procedure. As illustrated in Fig. \ref{fig:appendix-visualize-layout-control}, we observe that sampling with mask priors effectively controls the approximate position of objects.

\begin{figure}[t!]
\begin{minipage}{0.5\linewidth}
  \centering
  \includegraphics[width=0.9\linewidth]{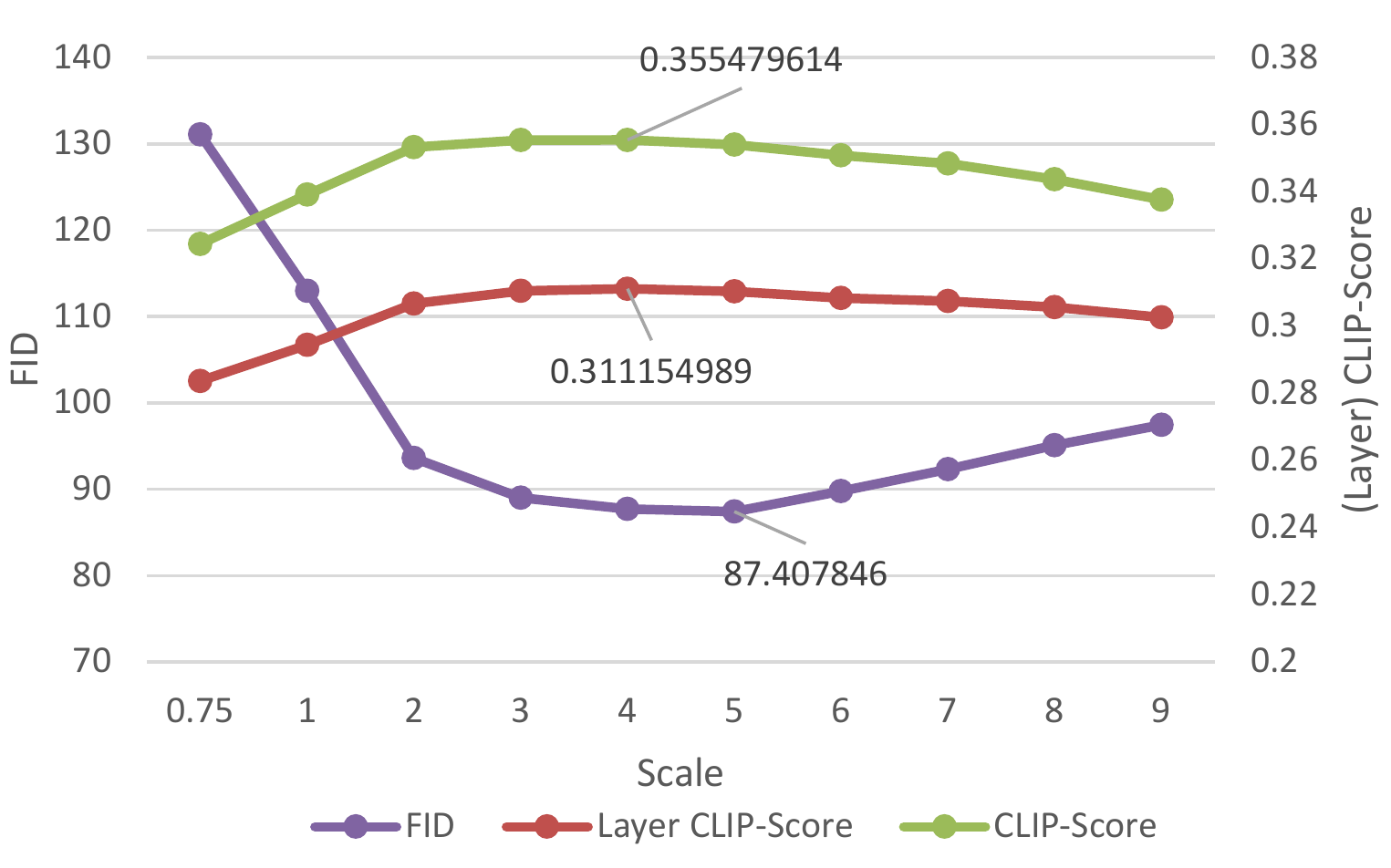}
  \caption{The effect of different scale of Self-Mask Guidance sampling. The best value of each metric is indicated in the figure.}
  \label{fig:appendix-visualize-curve-of-smg}
\end{minipage}%
~
\begin{minipage}{0.5\linewidth}
  \centering
  \includegraphics[width=\linewidth]{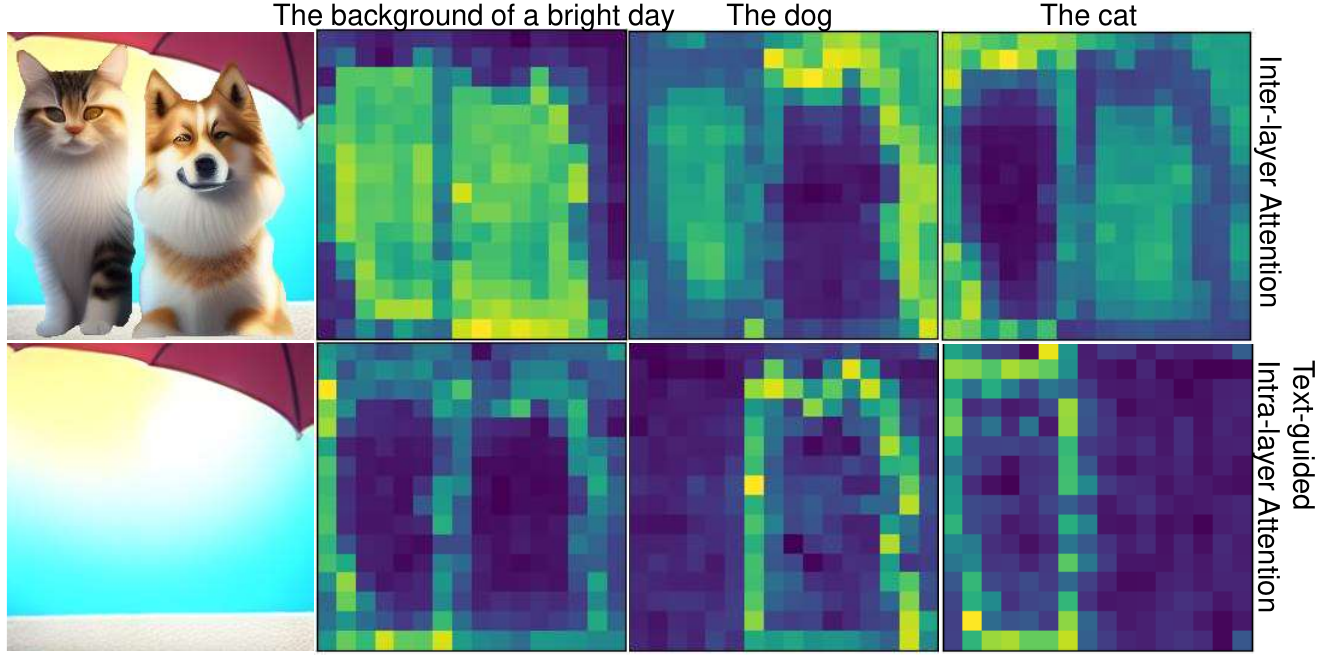}
  \caption{The visualization of attention map of the inter-layer attention and text-guided intra-layer attention.}
  \label{fig:appendix-visualize-attention-map}
\end{minipage}
\end{figure}

\begin{figure}[ht!]
\vspace{-1em}
\begin{center}
\includegraphics[width=0.9\linewidth]{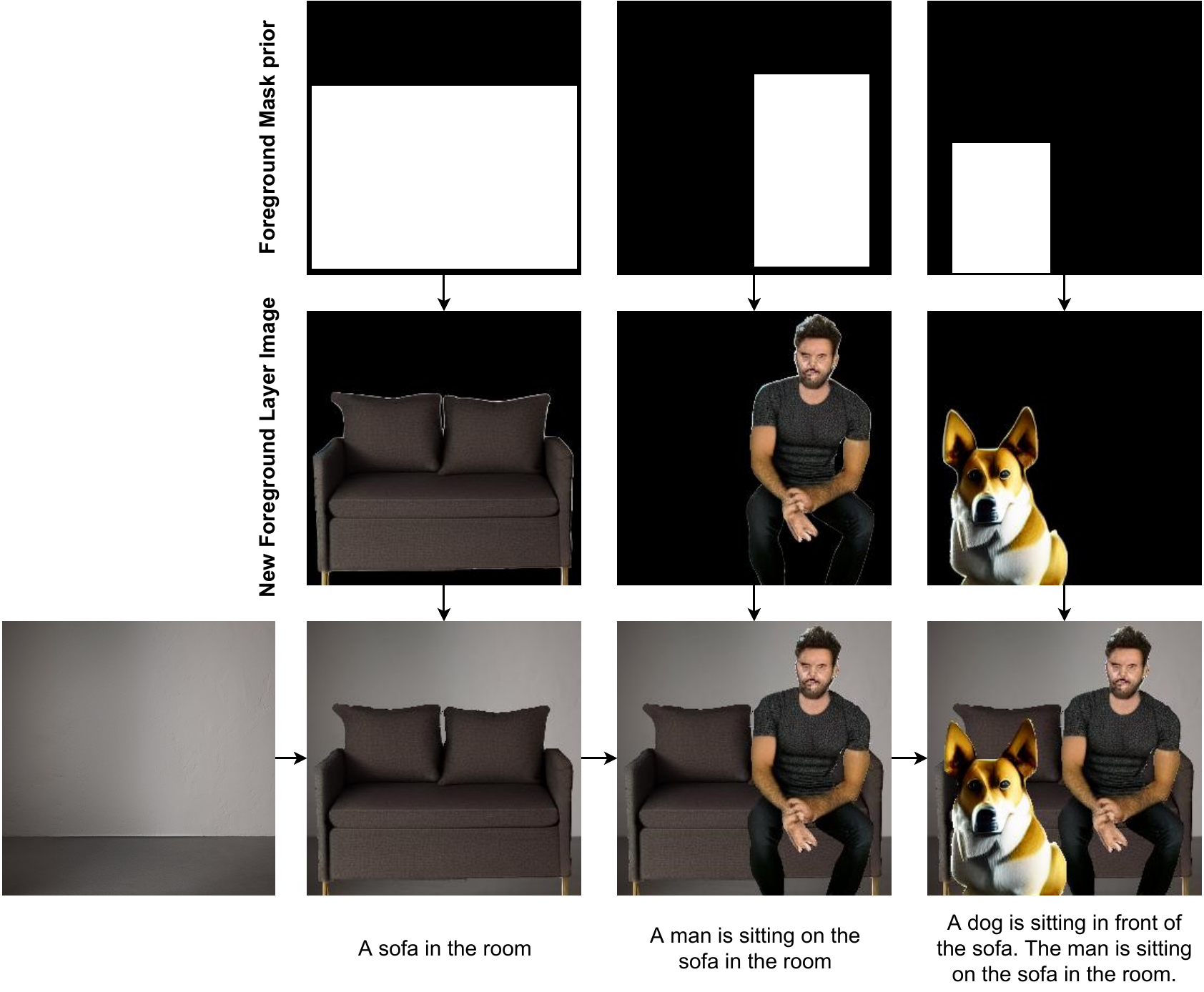}
\vspace{-1em}
\caption{An example to iterative generating multi-layered composable image. We achieve multi-layered composable image synthesis by merging the newly generated foreground layer into the background layer and iteratively performing foreground layer inpainting on two-layered images.}
\label{fig:appendix-visualize-expanding-layer-control}
\end{center}
\vspace{-3mm}
\end{figure}
\subsection{Iterative Generating Multi-layered Composable Images}
The multi-layered composable image synthesis can be achieved by repeatedly inpainting foreground layers while maintaining a two-layered image structure. Specifically, after generating a new foreground layer, we merge it with the original background layer to form a new background layer, then perform foreground layer inpainting again. As shown in Fig. \ref{fig:appendix-visualize-expanding-layer-control}, combined with mask prior, we successfully synthesize a four-layer composable image.

\section{Discussion on Self-Mask Guidance Sampling}
Table~\ref{tab:appendix-ablation-sag-smg} demonstrates the effectiveness of Self-Mask Guidance (SMG) sampling compared with the Self-Attention Guidance (SAG) sampling~\cite{hong2023improving}. In multi-layered composable image sampling, SMG can boost the sampling quality in both the degree of alignment and fidelity than the SAG. Moreover, as shown in Fig.~\ref{fig:appendix-visualize-curve-of-smg}, when using the classifier-free guidance scale of 3, increasing the scale of SMG significantly improves the quality of the generated images. Optimal performance across various metrics is achieved when employing scales of 4 or 5.


